\documentstyle[rotate]{article}

\newread\epsffilein    
\newif\ifepsffileok    
\newif\ifepsfbbfound   
\newif\ifepsfverbose   
\newdimen\epsfxsize    
\newdimen\epsfysize    
\newdimen\epsftsize    
\newdimen\epsfrsize    
\newdimen\epsftmp      
\newdimen\pspoints     
\def\epsfbox#1{\global\def\epsfllx{72}\global\def\epsflly{72}%
   \global\def\epsfurx{540}\global\def\epsfury{720}%
   \def\lbracket{[}\def\testit{#1}\ifx\testit\lbracket
   \let\next=\epsfgetlitbb\else\let\next=\epsfnormal\fi\next{#1}}%
\def\epsfgetlitbb#1#2 #3 #4 #5]#6{\epsfgrab #2 #3 #4 #5 .\\%
   \epsfsetgraph{#6}}%
\def\epsfnormal#1{\epsfgetbb{#1}\epsfsetgraph{#1}}%
\def\epsfgetbb#1{%
\openin\epsffilein=#1
\ifeof\epsffilein\errmessage{I couldn't open #1, will ignore it}\else
   {\epsffileoktrue \chardef\other=12
    \def\do##1{\catcode`##1=\other}\dospecials \catcode`\ =10
    \loop
       \read\epsffilein to \epsffileline
       \ifeof\epsffilein\epsffileokfalse\else
          \expandafter\epsfaux\epsffileline:. \\%
       \fi
   \ifepsffileok\repeat
   \ifepsfbbfound\else
    \ifepsfverbose\message{No bounding box comment in #1; using defaults}\fi\fi
   }\closein\epsffilein\fi}%
\def\epsfsetgraph#1{%
   \epsfrsize=\epsfury\pspoints
   \advance\epsfrsize by-\epsflly\pspoints
   \epsftsize=\epsfurx\pspoints
   \advance\epsftsize by-\epsfllx\pspoints
   \epsfxsize\epsfsize\epsftsize\epsfrsize
   \ifnum\epsfxsize=0 \ifnum\epsfysize=0
      \epsfxsize=\epsftsize \epsfysize=\epsfrsize
     \else\epsftmp=\epsftsize \divide\epsftmp\epsfrsize
       \epsfxsize=\epsfysize \multiply\epsfxsize\epsftmp
       \multiply\epsftmp\epsfrsize \advance\epsftsize-\epsftmp
       \epsftmp=\epsfysize
       \loop \advance\epsftsize\epsftsize \divide\epsftmp 2
       \ifnum\epsftmp>0
          \ifnum\epsftsize<\epsfrsize\else
             \advance\epsftsize-\epsfrsize \advance\epsfxsize\epsftmp \fi
       \repeat
     \fi
   \else\epsftmp=\epsfrsize \divide\epsftmp\epsftsize
     \epsfysize=\epsfxsize \multiply\epsfysize\epsftmp   
     \multiply\epsftmp\epsftsize \advance\epsfrsize-\epsftmp
     \epsftmp=\epsfxsize
     \loop \advance\epsfrsize\epsfrsize \divide\epsftmp 2
     \ifnum\epsftmp>0
        \ifnum\epsfrsize<\epsftsize\else
           \advance\epsfrsize-\epsftsize \advance\epsfysize\epsftmp \fi
     \repeat     
   \fi
   \ifepsfverbose\message{#1: width=\the\epsfxsize, height=\the\epsfysize}\fi
   \epsftmp=10\epsfxsize \divide\epsftmp\pspoints
   \vbox to\epsfysize{\vfil\hbox to\epsfxsize{%
      \includegraphics{#1}%
      \hfil}}%
\epsfxsize=0pt\epsfysize=0pt}%

{\catcode`\%=12 \global\let\epsfpercent=
\long\def\epsfaux#1#2:#3\\{\ifx#1\epsfpercent
   \def\testit{#2}\ifx\testit\epsfbblit
      \epsfgrab #3 . . . \\%
      \epsffileokfalse
      \global\epsfbbfoundtrue
   \fi\else\ifx#1\par\else\epsffileokfalse\fi\fi}%
\def\epsfgrab #1 #2 #3 #4 #5\\{%
   \global\def\epsfllx{#1}\ifx\epsfllx\empty
      \epsfgrab #2 #3 #4 #5 .\\\else
   \global\def\epsflly{#2}%
   \global\def\epsfurx{#3}\global\def\epsfury{#4}\fi}%
\def\epsfsize#1#2{\epsfxsize}

\newlength{\spatmplen}

\newfont{\df}{cmssbx10}

\newfont{\enormous}{cmr17 scaled\magstep5}

\newcommand\Tilde{$\sim$}

\newcommand{\sub}[1]{\(_{\mbox{\scriptsize #1}}\)}

\newcommand{\goto}{\(\rightarrow\)}

\newcommand{\defop}[2]{\newcommand#1{\mathop{\rm #2}}}
\defop{\argmax}{argmax}
\defop{\PR}{\mbox{\df P}}
\defop{\E}{\mbox{\df E}}

\defop{\Bin}{Bin}

\newbox\inviotabox
\setbox\inviotabox=\hbox{$\iota$}

\newcommand\BI{\begin{itemize}}
\newcommand\EI{\end{itemize}}
\newcommand\BE{\begin{enumerate}}
\newcommand\EE{\end{enumerate}}

\def\descriptionlabel#1{{\bf #1 \hfil}}
\def\description{\list{}{\itemindent 0ex \labelwidth 10ex \itemsep 2ex \leftmargin 14ex
       \let\makelabel\descriptionlabel}}

\newcounter{excount}
\newcounter{tmpa}
\newcommand{\px}{\theexcount}

\newcommand{\nx}{%
  \setcounter{tmpa}{\value{excount}}%
  \addtocounter{tmpa}{1}%
  \thetmpa%
}

\newcounter{subexcount}[excount]

\newenvironment{example}{
  \begin{list}{}{
    \itemindent 0ex \labelwidth 6ex \itemsep 1ex \leftmargin 7ex
    \refstepcounter{excount}
    \let\makelabel\descriptionlabel
  }
  \item[(\theexcount)\hfill]
}{\end{list}}

\newenvironment{noexample}[1]{
  \begin{list}{}{
    \itemindent 0ex \labelwidth 6ex \itemsep 1ex \leftmargin 7ex
    \let\makelabel\descriptionlabel
  }
  \item[#1]
}{\end{list}}

\newenvironment{subexamples}{
  \begin{list}{
    \alph{subexcount}.\hfill
  }{
    \itemindent 0ex \labelwidth 3ex \itemsep 1ex \leftmargin 4ex
    \usecounter{subexcount}
    \let\makelabel\descriptionlabel
  }
}{\end{list}}

\newcounter{exercise}

\newcounter{SpaEnumSaved}

\newcommand\spaPsScale{1}
\newcommand\psscale[1]{\def\spaPsScale{#1}}
\def\epsfsize#1#2{\spaPsScale#1}

\newcommand{\ps}[1]{
  \mbox{
    \setbox1=\hbox{\epsfbox{#1.ps}}
    \dimen1=\ht1\advance\dimen1 by -2ex
    \lower\dimen1\box1
  }
}

\bibliographystyle{plain}

\title{Stochastic Attribute-Value Grammars}
\author{Steven Abney \\ University of T\"ubingen}
\date{}

\def\normpsscale{\psscale{0.7}}

\newenvironment{mathexample}{
\begin{example}
\begin{math}
\begin{displaystyle}
}{
\end{displaystyle}
\end{math}
\end{example}
}

\begin{document}

\normpsscale
\newcommand{\diverg}[2]{D({#1}|\!|{#2})}
\newcommand{\diverge}[1]{\diverg{\tilde{p}}{#1}}
\newcommand{\qold}{q_{\mbox{\scriptsize old}}}

\maketitle

\begin{center}
http://www.sfs.nphil.uni-tuebingen.de/\Tilde abney/\\
abney@sfs.nphil.uni-tuebingen.de\\
Wilhelmstr.\ 113, 72074 T\"ubingen, Germany
\end{center}

\begin{abstract}
Probabilistic analogues of regular and context-free grammars are
well-known in computational linguistics, and currently the subject of
intensive research.  To date, however, no satisfactory probabilistic
analogue of attribute-value grammars has been proposed: previous
attempts have failed to define a correct parameter-estimation
algorithm.  

In the present paper, I define stochastic attribute-value grammars and
give a correct algorithm for estimating their parameters.  The 
estimation algorithm is adapted from Della Pietra, Della
Pietra, and Lafferty \cite{DellaPietra_95}.  To estimate model parameters,
it is necessary to compute the expectations of certain functions under
random fields.  In the application discussed by Della Pietra, Della
Pietra, and Lafferty (representing English
orthographic constraints), Gibbs sampling can be used to estimate the
needed expectations.  The fact that attribute-value grammars generate
constrained languages makes Gibbs sampling inapplicable, but I show
how a variant of Gibbs sampling, the Metropolis-Hastings algorithm,
can be used instead.
\end{abstract}

\section{Introduction}

Stochastic versions of regular grammars and context-free grammars have
received a great deal of attention in computational linguistics for
the last several years, and basic techniques of stochastic parsing and
parameter estimation have been known for decades.  However, regular
and context-free grammars are widely deemed linguistically inadequate;
standard grammars in computational linguistics are attribute-value
grammars of some variety.  Before the advent of statistical methods,
regular and context-free grammars were considered too inexpressive for
serious consideration, and even now the reliance on stochastic
versions of the less-expressive grammars is often seen as an expedient
necessitated by the lack of an adequate stochastic version of
attribute-value grammars.

Attempts have been made to extend stochastic models developed for the
regular and context-free cases to attribute-value grammars, but to
date without success.\footnote{I confine my discussion here to Brew
and Eisele because they aim to describe parametric models of probability
distributions over the languages of constraint-based grammars, and to
estimate the parameters of those models.  Other authors have assigned
weights or preferences to constraint-based grammars but not discussed
parameter estimation.  One approach of the latter sort that I find of
particular interest is that of Stefan Riezler
\cite{Riezler_96}, who describes a weighted logic for constraint-based
grammars that characterizes the languages of the grammars as fuzzy
sets.  This interpretation avoids the need for normalization that Brew
and Eisele face, though parameter estimation still remains to be addressed.}
Brew \cite{Brew_95} sketches a probabilistic
version of HPSG, but admits that his way of dealing with re-entrancies
in feature structures is problematic.  Eisele \cite{Eisele_94} attempts to translate stochastic
context-free techniques to constraint-based grammar by assigning
probabilities to SLD proof trees.  Both Brew and Eisele propose
associating weights with grammar-rule analogues (typed feature
structures in Brew's case; Horn clauses in Eisele's case) and setting
weights proportional to expected rule frequencies.  For want of a
standard term, I will call this the Expected Rule Frequency (ERF)
method.  Both propose using iterative reestimation of rule-frequency
expectations when dealing with incomplete data (unannotated corpora),
along the lines of the EM algorithm.

The attempt is ultimately unsuccessful.  The ERF method is provably
correct for the context-free case, but it fails in the presence of
context dependencies, as will be discussed below.  Both Brew and
Eisele recognize that applying the ERF method has deficiencies.
Eisele in particular identifies an important symptom that indicates that
something has gone amiss: the grammar induced by the EM algorithm
defines a probability distribution over trees that is not in
accordance with their frequency in the training corpus.  Moreover,
Eisele recognizes that this problem arises only where there are
context dependencies.  That such dependencies lead to
problems is not surprising, given the independence assumptions
underlying Eisele's model, but he is not able to explain why they
manifest themselves in the way they do, nor what can be done to
address the problem.

Now in fact solutions to the context-sensitivity problem have long
been known, and are the subject of continuing study, in the image
processing field and in related areas of statistics.  The models of
interest are known as random fields.  Random fields can be seen as a
generalization of Markov chains and stochastic branching processes.
Markov chains can be seen as stochastic versions of regular grammars
(Hidden Markov Models are in turn stochastic functions of Markov chains)
and random branching processes are stochastic versions of context-free
grammars.  The evolution of a Markov chain describes
a line, in which each stochastic choice depends only on the state at
the immediately preceding time-point.  The evolution of a random
branching process describes a tree in which a finite-state process may
spawn multiple child processes at the next time-step, but the number
of processes and their states depend only on the state of the unique
parent process at the preceding time-step.  In particular, stochastic
choices are {\em independent} of other choices at the same time-step:
each process evolves independently.  If we permit re-entrancies, that
is, if we permit processes to re-merge, we generally introduce
context-sensitivity.  In order to re-merge, processes must generally
be ``in synch,'' which is to say, they cannot evolve in complete
independence of one another.  Random fields are a particular class of
multi-dimensional random processes, that is, processes corresponding
to probability distributions over an arbitrary graph.
They were originally studied by
Gibbs, nearly a hundred years ago, as a model for statistical
mechanics, and the general family of probability distributions
involved is still known by his name.

To my knowledge, the first application of random fields to natural
language was by Mark et al.\ \cite{Mark_92}.  The problem of interest
was how to combine a stochastic context-free grammar with n-gram
language models.  The resulting structures, e.g., (\nx), obviously involve
re-entrancies and context-sensitivity.

\begin{example}
\ps{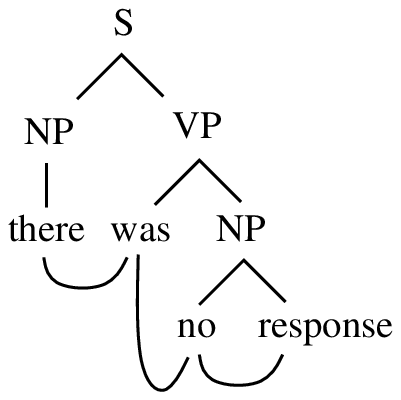}
\end{example}

\noindent It was clear at that time that a similar approach ought to
succeed for general attribute-value grammars, but the issue was not
pursued.

Recent work by Della Pietra, Della Pietra, and Lafferty
\cite{DellaPietra_95} (henceforth, DDL) also applies random fields to
natural language processing.  The application they consider is the
induction of English orthographic constraints---inducing a grammar of
possible English words.  The authors describe an algorithm for
selecting informative properties of words to construct a random
field, and for setting the parameters of the field optimally for a
given set of properties, to model an empirical word distribution.

The DDL algorithms require the computation of the expectations, under
random fields, of
certain characteristic functions.  In general,
computing these expectations involves summing over all configurations
(all possible character sequences, in the orthography application),
which is not possible when the configuration space is large.  Instead,
DDL use Gibbs sampling to estimate the needed expectations.

The orthography application cannot be immediately converted into a
means of equipping attribute-value grammars with probabilities.  Any
labelling of a finite linear graph\footnote{To be precise, DDL use
closed linear graphs---i.e., polygons.} with ASCII characters yields a
possible (though not necessarily probable) English word, and this
unconstrainedness is essential for the use of Gibbs sampling.  By
contrast, the set of dags admitted by an attribute-value grammar $G$ is
highly constrained---most of the time, relabelling a dag admitted by
$G$ does {\em not} yield a new dag admitted by $G$.  Gibbs sampling is
not applicable.  However, I will show that a variant of Gibbs
sampling, the Metropolis-Hastings algorithm, can be used.  Indeed, we
can use a random branching process much like Brew's or Eisele's to
supply the so-called proposal matrix for the Metropolis-Hastings
algorithm.

In this way, we can assign probabilities to the classes of dags
admitted by attribute-value grammars.  We can use these probabilities
to disambiguate sentences (by selecting the most-probable parse), and
we can give a parameter-estimation algorithm that is {\em correct,} in
the sense that, if we generate a training corpus of size $n$ from a
model $M$, and then estimate parameters from the training corpus to
yield a model-estimate $\hat{M}_n$, then $\hat{M}_n$ converges to $M$ as
$n \rightarrow \infty$.

\subsection*{Acknowledgements}

This work has greatly profited from the comments, criticism, and
suggestions of a number of people, including John Lafferty, Stanley
Peters, Hans Uszkoreit, and other members of the audience for talks I
gave at Saarbr\"ucken and T\"ubingen.  Michael Miller and Kevin Mark
introduced me to random fields as a way of dealing with
context-sensitivities in language, and I have been fascinated ever
since.  I would especially like to thank
Mark Light and Stefan Riezler for extended discussions of the issues
addressed here and helpful criticism on points of presentation.  All
responsibility for flaws and errors of course remains with me.

\section{Stochastic Context-Free Grammars}

Let us begin by examining stochastic context-free grammars and asking
why the ``obvious'' generalization to attribute-value grammars fails.
A point of terminology: I will use the term {\it grammar} to refer to
an unweighted grammar, be it a context-free grammar or attribute-value
grammar.  The combination of a grammar and weights (later, also
properties) I will refer to as a {\it model}.  (Occasionally I will
use {\it model} to refer to the weights themselves, or the probability distribution
they define.)

Throughout we will use the following stochastic context-free grammar
for illustrative purposes.  Let us call the underlying grammar $G_1$
and the grammar equipped with weights as shown, $M_1$:

\begin{example}\label{G1}
\begin{tabular}[t]{lll}
1. & S \goto\ A A & $\beta_1 = 1/2$ \\
2. & S \goto\ B   & $\beta_2 = 1/2$ \\[2ex]
3. & A \goto\ a   & $\beta_3 = 2/3$ \\
4. & A \goto\ b   & $\beta_4 = 1/3$ \\[2ex]
5. & B \goto\ a a  & $\beta_5 = 1/2$ \\
6. & B \goto\ b b  & $\beta_6 = 1/2$
\end{tabular}
\end{example}

\noindent The probability of a given tree is computed as the product
of probabilities of rules used in it.  For example:

\begin{example}\label{t1}
\ps{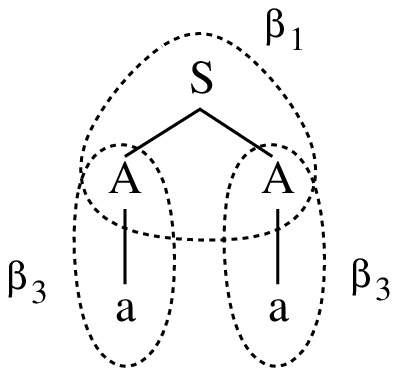}
\end{example}

\noindent Let $x$ be tree (\px) and let $q_1$ be the probability
distribution over trees defined by model $M_1$.  Then:

\begin{example}
\[
\begin{array}{lll}
q_1(x) &=& \beta_1 \cdot \beta_3 \cdot \beta_3 \\
     &=& \frac{1}{2} \cdot \frac{2}{3} \cdot \frac{2}{3} = \frac{2}{9}
\end{array}
\]
\end{example}

In parsing, we use the probability distribution $q_1(x)$ defined by
model $M_1$ to
disambiguate: the grammar assigns some set of trees $\{x_1, \ldots,
x_n\}$ to a sentence $\sigma$, and we choose that tree $x_i$ that has
greatest probability $q_1(x_i)$.  For example, $G_1$ assigns two parses
to the sentence {\it aa}: tree (\ref{t1}) above and tree (\nx):

\begin{example}
\ps{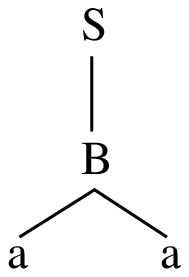}
\end{example}

\noindent The probability of tree (\ref{t1}) is 2/9, as we have seen.
The probability of tree (\px) is $\beta_2\beta_5 = 1/2 \cdot 1/2 =
1/4$.  Since $1/4 > 2/9$, a stochastic parser for $M_1$ should return
tree (\px) on input {\it aa}.

The issue of efficiently computing the most-probable parse for a given
sentence has been thoroughly addressed in the literature.  The standard
parsing techniques can be applied as is to the random-field models
to be discussed below, so I simply refer the reader to the literature.
Instead, I concentrate on parameter estimation, which for
attribute-value grammars cannot be accomplished by standard techniques.

By {\it parameter estimation} we mean determining values for the weights $\beta$.
In order for a stochastic grammar to be useful, we must be able to
compute the correct weights, where by {\it correct weights} we mean
the weights that best
account for a training corpus.  The degree to which a given set of
weights account for a training corpus is measured by the similarity between the
distribution $q_\beta(x)$ determined by the weights $\beta$ and the
distribution of trees $x$ in the training corpus.

\subsection{The Goodness of a Model}

The distribution determined by the training corpus is known as the
{\df empirical distribution}.  For example, suppose we have a training
corpus containing twelve trees of the following four types from $L(G_1)\/$:

\begin{example}\label{corpus1}
\ps{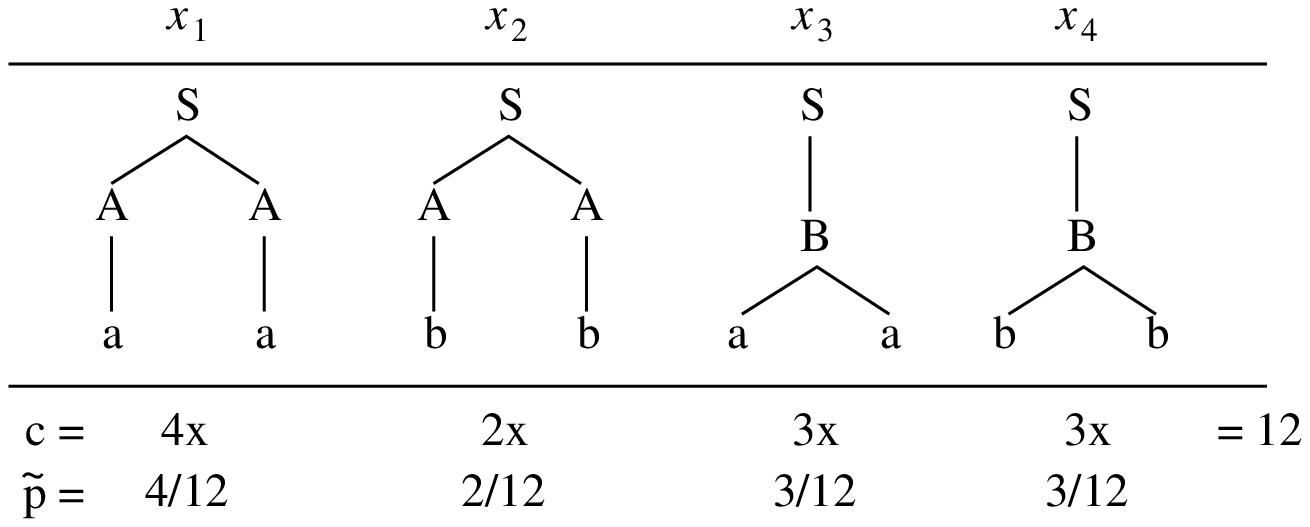}
\end{example}

\noindent If $c_i$ is the count of how often the $i$-th tree (type)
appears in the corpus, then

\[\tilde{p}(x_i) = \frac{c_i}{\sum_j c_j}\]

In comparing a distribution $q$ to the empirical distribution
$\tilde{p}$, we shall actually measure dissimilarity rather than
similarity.  Our measure for dissimilarity of distributions is the
Kullback-Leibler distance, defined as:

\begin{example}
\[\diverge{q} = \sum_{x} \tilde{p}(x) \ln \frac{\tilde{p}(x)}{q(x)}\]
\end{example}

\noindent The distance between $\tilde{p}$ and $q$ at point $x$ is the
log of the ratio of $\tilde{p}(x)$ to $q(x)$.  The overall distance
between $\tilde{p}$ and $q$ is the average distance, where the
averaging is over tree (tokens) in the corpus; i.e., point distances
$\ln \tilde{p}(x)/q(x)$ are weighted by $\tilde{p}(x)$ and summed.

For example, let $q_1$ be, as before, the distribution determined by
model $M_1$.  The following table shows $q_1$, $\tilde{p}$,
the ratio $q_1(x)/\tilde{p}(x)$, and the weighted point distance
$\tilde{p}(x)\ln(\tilde{p}(x)/q_1(x))$.  The sum of the fourth column is
the Kullback-Leibler distance $\diverge{q}$ between $\tilde{p}$
and $q_1$.  The third column contains $q_1(x)/\tilde{p}(x)$ rather than
$\tilde{p}(x)/q_1(x)$ so that one can see at a glance whether $q_1(x)$ is
too large ($q_1(x)/\tilde{p}(x) > 1$) or too small ($< 1$).

\begin{example}\label{divergeq1}
\begin{tabular}[t]{lllll}
      & $q_1$  & $\tilde{p}$ & $q_1/\tilde{p}$ & $\tilde{p}\ln(\tilde{p}/q_1)$ \\ \hline
$x_1$ & 2/9    & 1/3         & 0.67          & 0.14                      \\
$x_2$ & 1/18   & 1/6         & 0.33          & 0.18                      \\
$x_3$ & 1/4    & 1/4         & 1.00          & 0.00                      \\
$x_4$ & 1/4    & 1/4         & 1.00          & 0.00                      \\ \hline
      &        &             &               & {\bf 0.32}
\end{tabular}
\end{example}

\noindent The total distance $\diverge{q_1} = 0.32$.

One set of weights is better than another if its distance
from the empirical distribution is less.  For example, let us consider
a different set of weights for grammar $G_1$.  Let $M'$ be $G_1$ with weights
$(1/2, 1/2, 1/2, 1/2, 1/2, 1/2)$, and let $q'$ be the probability
distribution determined by $M'$.
Then the computation of the Kullback-Leibler distance is as follows:

\begin{example}
\begin{tabular}[t]{lllll}
      & $q'$  & $\tilde{p}$ & $q'/\tilde{p}$ & $\tilde{p}\ln(\tilde{p}/q')$ \\ \hline
$x_1$ & 1/8    & 1/3         & 0.38          & 0.33                      \\
$x_2$ & 1/8    & 1/6         & 0.75          & 0.05                      \\
$x_3$ & 1/4    & 1/4         & 1.00          & 0.00                      \\
$x_4$ & 1/4    & 1/4         & 1.00          & 0.00                      \\ \hline
      &        &             &               & {\bf 0.38}
\end{tabular}
\end{example}

\noindent The fit for $x_2$ improves, but that is more than offset by
a poorer fit for $x_1$.  The distribution $q_1$ is a better
distribution than $q'$, in the sense that $q_1$ is more similar
(less dissimilar) to the empirical distribution than $q'$
is.

This particular measure of goodness of a set of weights has a number
of nice properties.  For one thing, it is not hard to show that the
distribution closest to the empirical distribution is identically the
maximum likelihood distribution.

Another reason for adopting the definition of goodness in terms of
Kullback-Leibler distance is the following.  Suppose Nature secretly
chooses some set of weights $M$ for $G_1$.  These are the true
weights; they define the true distribution $q$.  Nature then generates
trees at random from $M$ in accordance with $q$.  Let $\tilde{p}_n$ be
the empirical distribution determined by the first $n$ trees that
Nature generates.  A
parameter-setting method must choose a model (a set of weights) $\hat{M}_n$ given
$\tilde{p}_n$, for each $n$.  A parameter-setting method is
correct if it converges to $M$, the true model.  The sequence of
hypotheses $\hat{M}_1, \hat{M}_2, \ldots$ defining distributions $\hat{q}_1, \hat{q}_2,
\ldots$ is said to converge to $M$ (defining distribution $q$) just in
case, for all tolerances $\epsilon$, there is some point $n$ such that
$\diverg{q}{\hat{q}_{n'}} < \epsilon$ for all $n' > n$.  It can be
shown that $\diverg{q}{\tilde{p}_n}$ converges to 0; that is,
$\lim_{n \rightarrow \infty} \tilde{p}_n = q$.  If a
parameter-setting method returns the model $\hat{M}_n$ that
minimizes $\diverg{\tilde{p}_n}{\hat{q}_n}$, then $\lim_{n \rightarrow
\infty} \hat{q}_n = \lim_{n \rightarrow \infty} \tilde{p}_n$, if the
limiting distribution for $\tilde{p}_n$ is generable by any model with
underlying grammar $G_1$.  Since $q$ is generable by such a grammar,
and $q$ is the limit distribution for $\tilde{p}_n$, it follows that
$q$ is also the limit distribution for $\hat{q}_n$, and the method is
correct.

Note that the model $\hat{M}$ that minimizes the distance
$\diverg{q}{\hat{q}}$ is $M$ itself, and $\diverg{q}{q} = 0$.  This
does not mean, however, that $\diverg{\tilde{p}_n}{\hat{q}_n} = 0$ for the model
minimizing $\diverg{\tilde{p}_n}{\hat{q}_n}$.  The empirical distributions
$\tilde{p}_n$ converge to $q$, but do not necessarily equal $q$.
Intuitively, the relative frequency of any given tree converges to its
true probability, but need not be precisely its true probability, even
in very large corpora.

\subsection{The ERF Method}

For stochastic context-free grammars, it can be shown that
the Expected Rule Frequency (ERF) method mentioned in the introduction always
yields the best model for a given training corpus.
To define the ERF method, we require a bit of terminology and notation.  With each rule $i$
in a stochastic context-free grammar is associated a weight $\beta_i$
and a function $f_i(x)$ that returns the number of times rule $i$ is
used in the derivation of tree $x$.  For example, consider tree
(\ref{t1}), repeated here as (\nx):

\begin{example}
\ps{fig01}
\end{example}

\noindent Rule 1 is used once and rule 3 is used twice; accordingly
$f_1(x) = 1$, $f_3(x) = 2$, and $f_i(x) = 0$ for $i \in \{2, 4, 5, 6\}$.

The expectation of a function over a probability space (for each $i$,
$f_i$ is such a function) is simply the average value of the function.
We use the notation $p[f]$ to represent the expectation of $f$ under
probability distribution $p$.  It is defined as:

\[p[f] = \sum_x p(x)f(x)\]

The ERF method
instructs us to choose the weight for rule $i$ proportional to the
average frequency of rule $i$ in the corpus.  That is:

\[\beta_i \propto \tilde{p}[f_i]\]

\noindent Algorithmically, we compute the expectation of each rule's
frequency, and normalize among rules with the same lefthand side.  For
example, consider corpus (\ref{corpus1}).  The expectation of each
rule frequency $f_i$ is a sum of terms $\tilde{p}(x)f_i(x)$.  These terms
are shown for each tree, in the following table.

\[
\begin{tabular}{llc|cc|cc|cc|}
     & &             & \rotate{S \goto\ A A} & \rotate{S \goto\ B} & 
\rotate{A \goto\ a} & \rotate{A \goto\ b} & \rotate{B \goto\ a a} & 
\rotate{B \goto\ b b} \\
     & & $\tilde{p}$ & $\tilde{p}f_1$ & $\tilde{p}f_2$ & $\tilde{p}f_3$ &
$\tilde{p}f_4$ & $\tilde{p}f_5$ & $\tilde{p}f_6$ \\ \hline
$x_1$ & {}[\sub{S} [\sub{A} a] [\sub{A} a]] & 1/3 & 1/3 &     & 2/3 &     &     & \\
$x_2$ & {}[\sub{S} [\sub{B} a a]] & 1/6 & 1/6 &     &     & 2/6 &     & \\
$x_3$ & {}[\sub{S} [\sub{A} b] [\sub{A} b]] & 1/4 &     & 1/4 &     &     & 1/4 & \\
$x_4$ & {}[\sub{S} [\sub{B} b b]] & 1/4 &     & 1/4 &     &     &     & 1/4 \\ \hline
\multicolumn{3}{r|}{$\tilde{p}[f] =$} &
              1/2 & 1/2 & 2/3 & 1/3 & 1/4 & 1/4 \\
\multicolumn{3}{r|}{$\beta =$} &
              1/2 & 1/2 & 2/3 & 1/3 & 1/2 & 1/2
\end{tabular}
\]

\noindent For example, in tree $x_1$, rule 1 is used once and rule 3
is used twice.  The empirical probability of $x_1$ is 1/3, so $x_1$'s
contribution to $\tilde{p}[f_1]$ is $1/3 \cdot 1$, and its
contribution to $\tilde{p}[f_3]$ is $1/3 \cdot 2$.  The weight
$\beta_i$ is obtained from $\tilde{p}[f_i]$ by normalizing among rules
with the same lefthand side.  For example, the expected rule
frequencies $\tilde{p}[f_1]$ and $\tilde{p}[f_2]$ of rules with
lefthand side S already sum to 1, so they are adopted without change
as $\beta_1$ and $\beta_2$.  On the other hand, the expected rule
frequencies $\tilde{p}[f_5]$ and $\tilde{p}[f_6]$ for rules with
lefthand side B sum to 1/2, not 1, so they are doubled to yield
weights $\beta_5$ and $\beta_6$.
It should be observed that the resulting weights are precisely the
weights of model $M_1$.

It can be proven that the ERF weights are the best weights for a given
grammar, in the sense that they define the distribution that is most
similar to the empirical distribution.  That is, if $\beta$ are the
ERF weights (for a given grammar), then $\diverge{q_\beta} <
\diverge{q_{\beta'}}$ for all sets of weights $\beta' \neq \beta$.

As noted earlier, one might expect the best weights to yield
$\diverge{q} = 0$, but such is
not the case.  We have just seen, for example, that the best
weights for grammar $G_1$ yield distribution $q_1$, yet
$\diverge{q_1} = 0.32 > 0$.  A close inspection of
the distance calculation (\ref{divergeq1}) reveals that $q_1$ is sometimes less than
$\tilde{p}$, but never greater than $\tilde{p}$.  Could we improve the
fit by increasing $q_1$?  For that matter, how can it be that $q_1$
is never greater than $\tilde{p}\/$?  As probability distributions,
$q_1$ and $\tilde{p}$ should have the same total mass, namely, 1.
Where is the missing mass for $q_1$?

The answer is of course that $q_1$ and $\tilde{p}$ are probability
distributions over $L(G)$, but not all of $L(G)$ appears in the
corpus.  Two trees are missing, and they account for the missing mass.
These two trees are:

\begin{example}
\ps{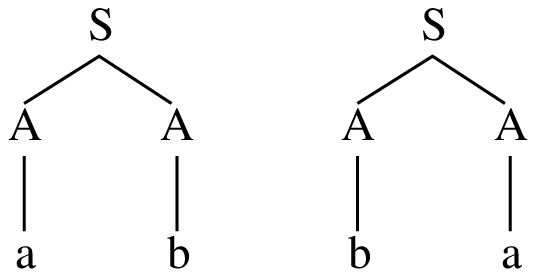}
\end{example}

\noindent Each of these trees have probability 0 according to
$\tilde{p}$ (hence they can be ignored in the distance calculation),
but probability $1/9$ according to $q_1$.

Intuitively, the problem is this.  The distribution $q_1$ assigns too
little weight to trees $x_1$ and $x_2$, and too much weight to the
trees of (\px); call them $x_5$ and $x_6$.  Yet exactly the same rules
are used in $x_5$ and $x_6$ as are used in $x_1$ and $x_2$.  Hence
there is no way to increase the weight for trees $x_1$ and
$x_2$, improving their fit to $\tilde{p}$, without simultaneously increasing
the weight for $x_5$ and $x_6$, making their fit to $\tilde{p}$ worse.
The distribution $q_1$ is the best compromise possible.

To say it another way, our assumption that the corpus was generated by
a context-free grammar means that any context dependencies in the
corpus must be accidental, the result of sampling noise.  There is
indeed a dependency in corpus (\ref{corpus1}): in the trees where
there are two A's, the A's always rewrite the same way.  If corpus
(\ref{corpus1}) was generated by a stochastic context-free
grammar, then this dependency is accidental.

This does not mean that the context-free assumption is wrong.  If we
generate twelve trees at random from $q_1$, it would not be too
surprising if we got corpus (\ref{corpus1}).  More extremely, if we
generate a random corpus of size 1 from $q_1$, it is quite impossible
for the resulting empirical distribution to match the distribution
$q_1$.  But as the corpus size increases, the fit between $\tilde{p}$
and $q_1$ becomes ever better.

\section{Attribute-Value Grammars}

But what if the dependency in corpus (\ref{corpus1}) is not
accidental?  What if we wish to adopt a grammar that imposes the
constraint that both A's rewrite the same way?
We can impose such a constraint by using an attribute-value grammar.
Consider the following grammar, in which rewrite rules are now
represented as feature structures.  Let us call this grammar $G_2$:

\begin{mathexample}\label{G2}
\begin{array}[t]{lll}
  \begin{array}[t]{ll}
    1.
    & 
    \left[\begin{array}{ll}
      S & \\
      1 & \left[\begin{array}{ll}
            A & \\
            1 & \fbox{1}
          \end{array}\right] \\
      2 & \left[\begin{array}{ll}
            A & \\
            1 & \fbox{1}
          \end{array}\right]
    \end{array}\right]
    \\
    2.
    &
    \left[\begin{array}{ll}
      S & \\
      1 & [B]
    \end{array}\right]
  \end{array}
  \begin{array}[t]{ll}
    3.
    &
    \left[\begin{array}{ll}
      A & \\
      1 & a
    \end{array}\right]
    \\
    4.
    &
    \left[\begin{array}{ll}
      A & \\
      1 & b
    \end{array}\right]
  \end{array}
  \begin{array}[t]{ll}
    5.
    &
    \left[\begin{array}{ll}
      B & \\
      1 & a
    \end{array}\right]
    \\
    6.
    &
    \left[\begin{array}{ll}
      B & \\
      1 & b
    \end{array}\right]
  \end{array}
\end{array}
\end{mathexample}

\noindent The language $L(G_2)$ is a set of dags, namely:

\begin{example}\label{L2}
\ps{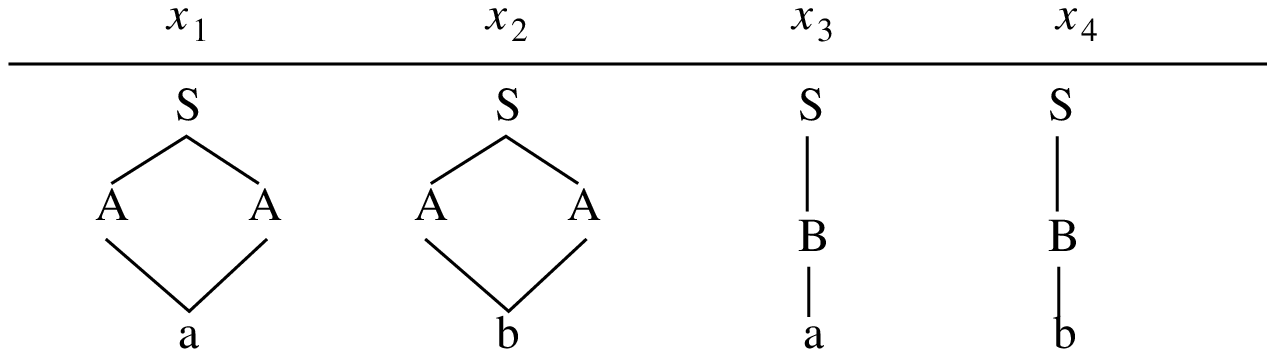}
\end{example}

\noindent (The edges of the dags should actually be labelled with 1's
and 2's, but I have suppressed the edge labels for the sake of
perspicuity.)

\subsection{AV Grammars and The ERF Method}

Now we face the question of how to attach probabilities to grammar
$G_2$.  The approach followed by Brew and Eisele is basically as
follows.\footnote{To be precise, neither Brew nor Eisele adopt the
attribute-value framework discussed here, but the approaches they take
in the related frameworks they do adopt are clearly analogous to the
one I describe here.}  Associate a weight with each of the
six ``rules'' of grammar $G_2$.  For example, let $M_2$ be the model
consisting of $G_2$ plus weights
$(\beta_1, \ldots, \beta_6) = (1/2, 1/2, 2/3, 1/3, 1/2, 1/2)$.
The weight assigned to a tree $x$ is then
(as before) the product of the weights of the rules used in $x$.  For
example, the weight $\breve{q}_2(x_1)$\footnote{The reason for the
`$\breve{\quad}$' will be made clear shortly.} assigned to tree $x_1$
of (\ref{L2}) is $2/9$, computed as follows:

\begin{example}
\ps{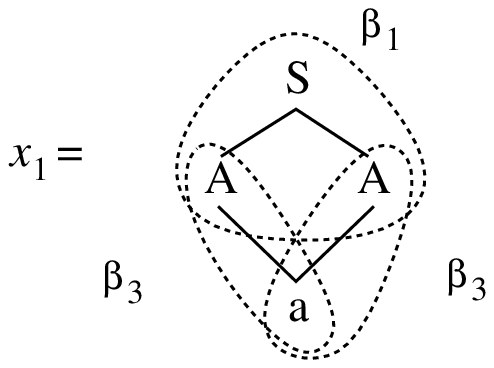}
\end{example}

\noindent Rule 1 is used once and rule 3 is used twice; hence
$\breve{q}_2(x_1) = \beta_1 \beta_3 \beta_3 = 1/2 \cdot 2/3 \cdot 2/3 =
2/9$.

Observe that $\breve{q}_2(x_1) = \beta_1 \beta_3^2$, which is to say,
$\beta_1^{f_1(x_1)}\beta_3^{f_3(x_1)}$.  Moreover, since $\beta^0 =
1$, it does not hurt to include additional factors
$\beta_i^{f_i(x_1)}$ for those $i$ where $f_i(x_1) = 0$.  That is, we
can define $\breve{q}_\beta$ corresponding to weights $\beta =
(\beta_1, \ldots, \beta_n)$ generally as:

\[\breve{q}_\beta(x) = \prod_{i=1}^{n} \beta_i^{f_i(x)}\]

Now let us consider how to estimate weights.
Brew and Eisele propose using the ERF method,
as in the context-free case.  To be sure, Brew and Eisele are more concerned
about the case in which the training corpus consists of sentences
alone, rather than parses (dags), and they concentrate on the application
of the EM algorithm to estimate rule-frequency expectations in the
absence of complete information.  But their basic method is the ERF
method: rule weights $\beta_i$ are set in accordance with the formula
$\beta_i \propto \tilde{p}[f_i]$, under the constraint that the
weights for rules with the same lefthand side sum to 1.  The EM
algorithm enters the picture only as a means of estimating
$\tilde{p}[f_i]$ when it cannot be determined by simple counting.

To illustrate, let us assume a corpus distribution for the dags
(\ref{L2}) analogous to the distribution in (\ref{corpus1}):

\begin{mathexample}\label{corpus2}
\begin{array}{lllll}
            & x_1 & x_2 & x_3 & x_4 \\
\tilde{p} = & 1/3 & 1/6 & 1/4 & 1/4
\end{array}
\end{mathexample}

\noindent Using the ERF method, we estimate rule weights as follows:

\begin{example}\label{erf}
\begin{tabular}{cc|cc|cc|cc|}
      & $\tilde{p}$ & $\tilde{p}f_1$ & $\tilde{p}f_2$ & $\tilde{p}f_3$ &
$\tilde{p}f_4$ & $\tilde{p}f_5$ & $\tilde{p}f_6$ \\ \hline
$x_1$ & 1/3 & 1/3 &     & 2/3 &     &     & \\
$x_2$ & 1/6 & 1/6 &     &     & 2/6 &     & \\
$x_3$ & 1/4 &     & 1/4 &     &     & 1/4 & \\
$x_4$ & 1/4 &     & 1/4 &     &     &     & 1/4 \\ \hline
\multicolumn{2}{r|}{$\tilde{p}[f] =$} &
              1/2 & 1/2 & 2/3 & 1/3 & 1/4 & 1/4 \\
\multicolumn{2}{r|}{$\beta =$} &
              1/2 & 1/2 & 2/3 & 1/3 & 1/2 & 1/2
\end{tabular}
\end{example}

\noindent This table is identical to the one given earlier in the
context-free case.  We arrive at the same weights we considered above
for the AV grammar $G_2$, yielding the distribution $\breve{q}_2$.

\subsection{Why the ERF Method Fails}

But at this point a problem arises: $\breve{q}_2$ is not a probability
distribution.  Unlike in the context-free case, the four trees in
(\ref{L2}) constitute the entirety of $L(G)$.  This time, there are no
missing trees to account for the missing probability mass.  There is
an obvious ``fix'' for this problem, as Brew and Eisele observe: we
can simply normalize $\breve{q}_2$.  (This, by the way, is the reason
for the `$\breve{\quad}$' in `$\breve{q}_2$'---it is meant to indicate
that $\breve{q}_2$ is an ``unnormalized'' probability distribution.)
That is, for the AV-grammar case, we must define the distribution $q_\beta$
corresponding to the weights $\beta$ as:

\[q_\beta(x) = \frac{1}{Z} \breve{q}_\beta(x)\]

\noindent where $Z$ is a normalizing constant defined as:

\[Z = \sum_{y \in L(G)} \breve{q}_\beta(y)\]

\noindent In particular, for the ERF weights given in (\ref{erf}), we
have $Z = 2/9 + 1/18 + 1/4 + 1/4 = 7/9$.  Dividing $\breve{q}_2$ by 7/9
yields the ERF distribution:

\begin{example}
\begin{tabular}{lllll}
         & $x_1$ & $x_2$ & $x_3$ & $x_4$ \\
$q_2(x) =$ & 2/7 & 1/14 & 9/28 & 9/28
\end{tabular}
\end{example}

On the face of it, then, we can transplant the methods we used in the
context-free case to the AV case and the only problem that arises
($\breve{q}_2$ not summing to 1) has an obvious fix (normalization).
However, something has actually gone very wrong.  The theorem
according to which the ERF method yields the best weights makes
certain assumptions that we inadvertently violated by changing $L(G)$
and re-apportioning probability via normalization.
In point of fact, we can
easily see that the ERF weights (\ref{erf}) are {\em not} the best
weights for our example grammar.  Consider the alternative model
$M*$ given in (\nx), defining probability distribution $q*$:

\begin{example}
\begin{tabular}{llllll}
[S A A] & [S B] & [A a a] & [A b b] & [B a a] & [B b b] \\
$\beta_1 =$ & $\beta_2 =$ & $\beta_3 =$ & $\beta_4 =$ & $\beta_5 =$ &
$\beta_6 =$ \\
$\frac{3+2\sqrt{2}}{6 + 2\sqrt{2}}$ & $\frac{3}{6 + 2\sqrt{2}}$ &
$\frac{\sqrt{2}}{1+\sqrt{2}}$ & $\frac{1}{1+\sqrt{2}}$ & $\frac{1}{2}$ & $\frac{1}{2}$
\end{tabular}
\end{example}

\noindent These weights are proper, in the sense that weights for
rules with the same lefthand side sum to one.  The reader can verify
that $\breve{q}*$ sums to $Z = \frac{3+\sqrt{2}}{3}$ and that
$q*$ is:

\psscale{0.3}
\begin{example}
\begin{tabular}[t]{lcccc}
         & \ps{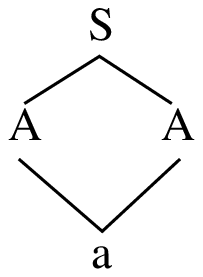} & \ps{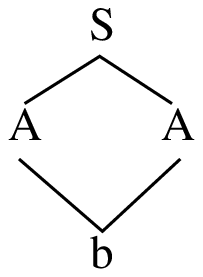} & \ps{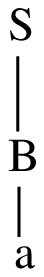} & \ps{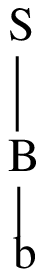} \\
         & $x_1$ & $x_2$ & $x_3$ & $x_4$ \\
${q*}(x) =$ & 1/3 & 1/6 & 1/4 & 1/4
\end{tabular}
\end{example}
\normpsscale

\noindent That is, $q* = \tilde{p}$.  Comparing $q_2$ (the ERF distribution)
and $q*$ to $\tilde{p}$, we observe that $\diverge{q_2}
= 0.07$ but $\diverge{q*} = 0$.  

In short, in the AV case, the ERF weights do {\em not} yield the best
weights.  This means that the ERF method does {\em not} converge to
the correct weights as the corpus size increases.  If there are
genuine dependencies in the grammar, the ERF method converges
systematically to the wrong weights.
Fortunately, there are methods that do converge to the right weights.
These are methods that have been developed for random fields.

\section{Random Fields}

A random field defines a probability distribution over a set of
labelled graphs $\Omega$ called {\it configurations}.  In our case,
the configurations are the dags generated by the grammar, i.e.,
$\Omega = L(G)$.\footnote{Those familiar with random fields will
recognize that identifying configurations with the dags of $L(G)$ is
not entirely unproblematic.  For one thing, configurations are
standardly taken to be labelings over a fixed graph, not graphs with
varying topologies.  For another thing, the configuration space is
standardly taken to be finite, not countably infinite, as $L(G)$ may
be.  These issues will be dealt with in the course of discussion.} The
weight assigned to a
configuration is the product of the weights assigned to configuration
properties.\footnote{The standard term in the random-fields literature
is {\it feature}; I use the term {\it property} to avoid confusion
with {\it feature} in the sense of an attribute plus value.} That is:

\[\breve{q}(x) = \prod_i \beta_i^{f_i(x)}\]

\noindent where $\beta_i$ is the weight for property $i$ and $f_i(x)$
is the frequency of occurence of property $i$ in configuration $x$.
The probability of a configuration is proportional to its weight, and is
obtained by normalizing the weight distribution.  That is:

\[\begin{array}{lll}
q(x) &=& \frac{1}{Z}\breve{q}(x) \\[1.5ex]
Z &=& \sum_{y \in \Omega} \breve{q}(y)
\end{array}\]

If we identify properties of a configuration with the rules used in
it, the random field model is almost identical to the model we
considered in the previous section.  There are two important
differences.  First, we no longer require weights to sum to
one for rules with the same lefthand side.  Second, we no longer
require properties to be identical to the rules of the grammar.  We
use the grammar to define the set of configurations $\Omega = L(G)$,
but give ourselves more flexibility in choosing the properties of dags
we would like to use to define the probability distribution over
$L(G)$.

Let us consider an example.  Let us continue to assume grammar
$G_2$ generating language (\ref{L2}), and let us continue to
assume the empirical distribution (\ref{corpus2}).  But now rather
than taking rule applications---local trees---to be properties, let us
adopt the following two properties:

\begin{example}
\ps{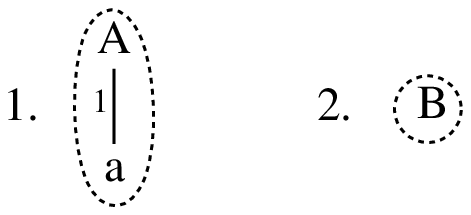}
\end{example}

\noindent For purpose of illustration, take property 1 to have weight
$\beta_1 = \sqrt{2}$ and property 2 to have weight $\beta_2 = 3/2$.
The functions $f_1$ and $f_2$ represent the frequencies of properties
1 and 2, respectively:

\psscale{0.5}
\begin{example}
\begin{tabular}[t]{rccccl}
                 & \ps{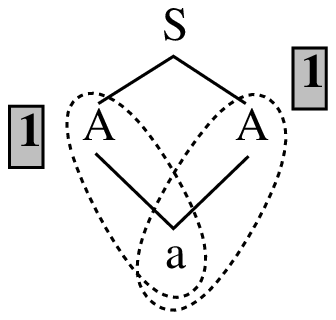} & \ps{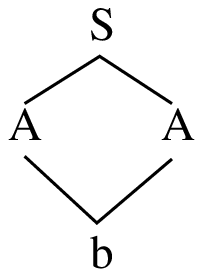} & \ps{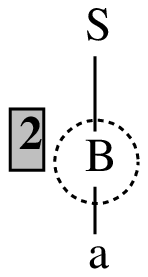} & \ps{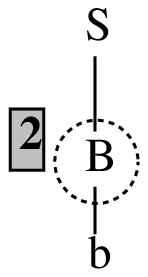} & \\
$f_1 =$       &     2      &      0     &    0       &      0     & \\
$f_2 =$       &     0      &      0     &    1       &      1     & \\
$\breve{q} =$ & $\sqrt{2}\cdot\sqrt{2}$ & 1 & 3/2    & 3/2        & $Z = 6$ \\
$q =$         & 2/6        &  1/6       & (3/2)/6      & (3/2)/6 \\
     =           & 1/3        & 1/6        & 1/4        & 1/4
\end{tabular}
\end{example}
\normpsscale

\noindent In short, we are able to exactly recreate the empirical
distribution using fewer properties than before.  Intuitively, we need
only use as many properties as are necessary to distinguish among
trees that have different empirical probabilities.

This added flexibility is welcome, but it does make parameter
estimation more involved.  Now we must not only choose values for
weights, we must also choose the properties that weights are to be
associated with.  We would like to do both in a way that permits us to
find the best model, in the sense of the model that minimizes the
Kullback-Leibler distance with respect to the empirical distribution.
Methods for doing both are given in a recent paper by Della Pietra,
Della Pietra, and Lafferty \cite{DellaPietra_95}.

\section{Field Induction}

In outline, the DDL algorithm is as follows:

\begin{enumerate}
\item Start ($t=0$) with the null field (no properties).

\item {\bf Property Selection.} Consider every property that might be
added to the field $q_t$ and choose the best one.

\item {\bf Weight Adjustment.}  Readjust weights for all properties.  The result
is a new field $q_{t+1}$.

\item Iterate until the field cannot be improved.
\end{enumerate}

One has a great deal of flexibility in defining the space of
properties.  For the sake of concreteness, let us take properties to
be labelled subdags.  In step 2 of the algorithm we do not consider
every conceivable labelled subdag (there are simply too many of
them), but only the atomic (i.e., single-node) subdags and those
complex subdags that can be constructed by combining properties
already in the field or by combining a property in the field with some
atomic property.

In our running example, the atomic properties are:

\begin{example}
\ps{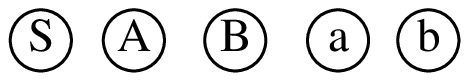}
\end{example}

\noindent Properties can be combined by adding connecting arcs.  For
example:

\psscale{0.5}
\begin{example}
\ps{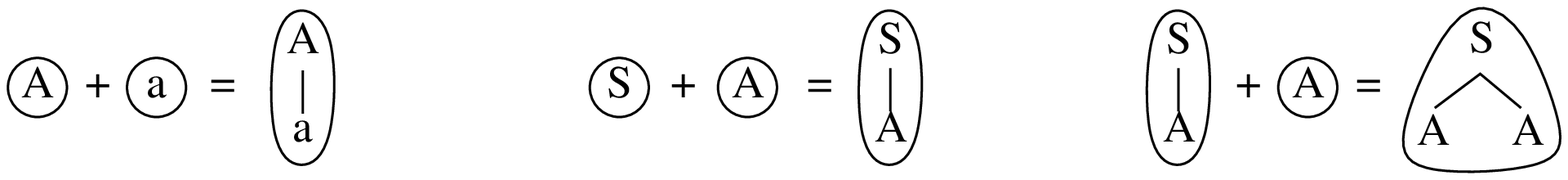}
\end{example}
\normpsscale

\subsection{The Null Field}

Field induction begins with the null field.  With the corpus we have
been assuming, the null field takes the following form.

\psscale{0.5}
\begin{example}
\begin{tabular}[t]{lccccl}
                 & \ps{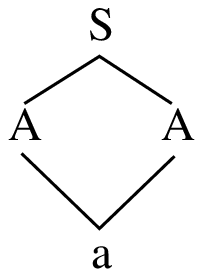} & \ps{fig08} & \ps{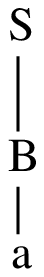} & \ps{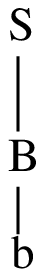} & \\
$\breve{q}(x) =$ & 1          & 1          & 1          & 1          & $Z = 4$ \\
$q(x) =$         & 1/4        &  1/4       & 1/4        & 1/4
\end{tabular}
\end{example}
\normpsscale

\noindent No dag $x$ has any features, so $\breve{q}(x) = \prod_i
\beta_i^{f_i(x)}$ is a product of zero terms, and hence has value 1.
As a result, $q$ is the uniform distribution.  The Kullback-Leibler
distance $\diverge{q}$ is 0.03.  The aim of property selection is to
choose a property that reduces this distance as much as possible.

The astute reader will note that there is a problem with the null
field if $L(G)$ is infinite.  Namely, it is not possible to have a
uniform distribution over an infinite set.  If each dag in an infinite
set of dags is assigned a constant nonzero probability $\epsilon$,
then the total probability is infinite, no matter how small $\epsilon$
is.  There are a couple of ways of dealing with the problem.  The
approach that DDL adopt is to assume a consistent prior distribution $p(k)$
over graph sizes $k$, and a family of random fields $q_k$ representing
the conditional probability $q(x|k)$; the probability of a tree is
then $p(k)q(x|k)$.  All the random fields have the same properties and
weights, differing only in their normalizing constants.

I will take a slightly different approach here.  Let us adopt an
initial distribution like that proposed by Brew and Eisele.  There is
a natural correspondence between AV grammars and CFG's, a
correspondence that we implicitly adopted in earlier discussion.
We assume that the rules of an AV grammar are typed feature
structures in which all types (of toplevel feature structures) are
disjoint.  Types correspond to categories in a CFG, and the righthand
side of the CF analogue of rule $r$ is the list of types of immediate
constituents of $r$, viewed as a feature structure.  For example, the
AV grammar $G_2$ has corresponding CF grammar $G_1$.

In this framework, a model consists of: (1)~An AV grammar $G$ whose purpose
is to define a set of dags $L(G)$.  (2)~An SCFG $H$ derived from $G$,
with weights $\theta$, defining a distribution $\breve{p}(d)$ over
derivations $d$.  There is a unique derivation corresponding to each
dag in $L(G)$, but some derivations correspond to no well-formed
dag---intuitively, some derivations lead to unification failures.
Discarding the bad derivations and renormalizing yields the initial
distribution $p(x)$ over dags $L(G)$.
(3)~A set of properties $f$ with weights $\beta$, to define the final
distribution $q(x) = \frac{1}{Z} \prod_i \beta_i^{f_i(x)} p(x)$.

There are a couple possible choices of weights $\theta$ for the initial
distribution.  The easiest approach would be to adopt the ERF weights.
Field induction would then be a way of adding context-sensitivities to
the ERF distribution.  An alternative would be to adopt
maximum-entropy weights.  The intuitive reason for adopting the
uniform distribution (in the finite case) is that it distinguishes
dags in $L(G)$ from dags not in $L(G)$, but otherwise makes no
assumptions about the distribution.  The uniform distribution
maximizes entropy over a finite set.  Maximizing entropy is more
generally applicable, however, and can be applied to infinite sets as
well.  Maximum entropy distributions for context-free languages are
discussed in a paper by Miller and O'Sullivan \cite{Miller_90a}, though a
number of technical questions arise that I do not wish to pursue
here.

\subsection{Property Selection}

At each iteration, we select a new property $f$ by considering all atomic
properties and all complex properties that can be constructed from
properties already in the field.  Holding the weights constant for all
old properties in the field, we choose the best weight $\beta$ for $f$
(how $\beta$ is chosen will be discussed shortly), yielding a new
distribution $q_f = q_{f,\beta}$.  The score for property $f$ is the
reduction it permits in $\diverge{q_{\mbox{\scriptsize old}}}$, where
$q_{\mbox{\scriptsize old}}$ is the old field.
That is, the score for $f$ is $\diverge{\qold} - \diverge{q_f}$.  We
compute the score for each candidate property and add to the field
that property with the highest score.

To illustrate, consider the two atomic properties `a' and `B'.  Given
the null field as old field, the best weight for `a' is $\beta = 7/5$,
and the best weight for `B' is $\beta = 1$.  This yields $q$ and
$\diverge{f}$ as follows:

\psscale{0.5}
\begin{example}
\begin{tabular}[t]{lccccl}
                 & \ps{fig19} & \ps{fig08} & \ps{fig20} & \ps{fig21} & \\
$\tilde{p}$     & 1/3        & 1/6        & 1/4        & 1/4 & \\[2ex]
$\breve{q}_a$ & 7/5        & 1        & 7/5          & 1          & $Z = 24/5$ \\
$q_a$         & 7/24        &  5/24     & 7/24       & 5/24       &\\
$\tilde{p} \ln \frac{\tilde{p}}{q_a}$
                 & 0.04       & $-0.04$    & $-0.04$    & $0.05$     &
$D = 0.01$ \\[2ex]
$\breve{q}_B$ & 1        & 1        & 1          & 1          & $Z = 4$ \\
$q_B$         & 1/4        &  1/4     & 1/4       & 1/4       &\\
$\tilde{p} \ln \frac{\tilde{p}}{q_B}$
                 & 0.10       &  $-0.07$    & 0    & $0$     &
$D = 0.03$
\end{tabular}
\end{example}
\normpsscale

\noindent The better property is `a', and `a' would be added to the
field if these were the only two choices.

Intuitively, `a' is better than `B' because `a' permits us to
distinguish the set $\{x_1,x_3\}$ from the set $\{x_2,x_4\}$; the
empirical probability of the former is $1/3 + 1/4 = 7/12$ whereas the
empirical probability of the latter is $5/12$.  Distinguishing these
sets permits us to model the empirical distribution better (since
the old field assigns them equal probability, counter to the empirical
distribution).  By contrast, the property `B' distinguishes the set
$\{x_1,x_2\}$ from $\{x_3,x_4\}$.  The empirical probability of the
former is $1/3 + 1/6 = 1/2$ and the empirical probability of the
latter is also $1/2$.  The old field models these probabilities
exactly correctly, so making the distinction does not permit us to
improve on the old field.  As a result, the best weight we can choose
for `B' is 1, which is equivalent to not having the property `B' at
all.

\subsection{Selecting the Initial Weight}

DDL show that there is a unique weight that maximizes the score for a
new property $f$ (provided that the score for $f$ is not constant for
all weights), and that the maximizing weight is the solution to the
equation

\begin{mathexample}\label{eq1}
q_{f,\beta}[f] = \tilde{p}[f]
\end{mathexample}

\noindent in the single unknown $\beta$.  Intuitively, we choose the
weight such that the expectation of $f$ under the resulting new field
is equal to its empirical expectation.

Solving equation (\ref{eq1}) for $\beta$ is easy if $L(G)$ is small enough
to enumerate.  Then the sum over $L(G)$ that is implicit in $q_{f,\beta}[f]$
can be expanded out, and solving for $\beta$ is simply a matter of
arithmetic.  Things are a bit trickier if $L(G)$ is too large to
enumerate.  DDL show that we can solve equation (\ref{eq1}) if we can
estimate $q_{\mbox{\scriptsize old}}[f = k]$ for $k$ from 0 to the maximum possible value for $f$.

We can estimate $q_{\mbox{\scriptsize old}}[f = k]$ by means of {\df random sampling}.  The
idea is actually rather simple: to estimate how often the property
appears in ``the average dag'', we generate a representative mini-corpus
from the distribution $q_{\mbox{\scriptsize old}}$ and count.  That is, we generate dags at
random in such a way that the relative frequency of dag $x$ is
$q_{\mbox{\scriptsize old}}(x)$ (in the limit), and we count how often the property of
interest appears in dags in our generated mini-corpus.

The application that DDL consider is the induction of English
orthographic constraints---inducing a field that assigns high
probability to ``English-sounding'' words and low probability to
non-English-sounding words.  For this application, Gibbs sampling is
appropriate.  Gibbs sampling does not work for the application to AV
grammars, however.  Fortunately, there is an alternative random
sampling method we can use: Metropolis-Hastings sampling.  We will
discuss the issue in some detail shortly.

\subsection{Readjusting Weights}

When a new property is added to the field, the best value for its
initial weight is chosen, but the weights for the old properties are
held constant.  In general, however, adding the new property may make
it necessary to readjust weights for all properties.  The second half
of the DDL algorithm involves finding the best set of weights for a
given set of properties.

The method is very similar to the method for selecting the initial
weight for a new property.  Let $(\gamma_1, \ldots, \gamma_n)$ be the
old weights for the properties.  Consider the equation

\begin{mathexample}\label{eq2}
q_\gamma[\beta_i^{f\#}f_i] = \tilde{p}[f_i]
\end{mathexample}

\noindent where $f\#(x) = \sum_i f_i(x)$ is the total number of
properties of dag $x$.  Without going into exactly why $f_i$ is weighted
as it is on the lefthand side, the idea is the same as before: we
want to adjust $\beta_i$ so that the average number of instances of
property $f_i$ according to the model matches the average number of
instances of property $f_i$ in dags in the corpus.

If the weights $\gamma_1, \ldots, \gamma_n$ are not already as good as
they can be, solving equation (\ref{eq2}) for $\beta_i$ (for each $i$)
is guaranteed to improve the weights, but it does not necessarily
immediately yield the globally best weights.  We can obtain the
globally best weights by iterating.  Set $\gamma_i \leftarrow
\beta_i$, for all $i$, and solve equation (\ref{eq2}) again.  Repeat until the
weights no longer change.

As with equation (\ref{eq1}), solving equation (\ref{eq2}) is
straightforward if $L(G)$ is small enough to enumerate, but not if
$L(G)$ is large.  In that case, we must use random sampling.
We generate a representative mini-corpus and estimate expectations by
counting in the mini-corpus.

\subsection{Random Sampling}

We have seen that random sampling is necessary both to set the initial
weight for properties under consideration and to adjust all weights
after a new property is adopted.  Random sampling involves creating a
corpus that is representative of a given model distribution $q(x)$.
To take a very simple example, a fair coin can be seen as a method for
sampling from the distribution $q$ such that $q(H) = 1/2$, $q(T) =
1/2$.  Saying that a corpus is representative is actually not a
comment about the corpus itself but the method by which it was
generated: a corpus representative of distribution $q$ is one
generated by a process that samples from $q$.  Saying that a process $M$
samples from $q$ is to say that the empirical distributions of corpora
generated by $M$ converge to $q$ in the limit.  For example, if we
flip a fair coin once, the resulting empirical distribution over
$(H,T)$ is either $(1,0)$ or $(0,1)$, not the fair-coin distribution
$(1/2,1/2)$.  But as we take larger and larger corpora, the resulting
empirical distributions converge to $(1/2,1/2)$.

One of the advantages of SCFGs, that is lost when we go to random
fields, is that there is a transparent relationship between an SCFG
defining a distribution $q$ and a sampler for $q$.  We can sample from
the distribution defined by an SCFG as follows.  Consider the grammar
(\ref{G1}), repeated here as (\nx):

\begin{example}
\begin{tabular}[t]{lll}
1. & S \goto\ A A & $\beta_1 = 1/2$ \\
2. & S \goto\ B   & $\beta_2 = 1/2$ \\[2ex]
3. & A \goto\ a   & $\beta_3 = 2/3$ \\
4. & A \goto\ b   & $\beta_4 = 1/3$ \\[2ex]
5. & B \goto\ a a   & $\beta_5 = 1/2$ \\
6. & B \goto\ b b   & $\beta_6 = 1/2$
\end{tabular}
\end{example}

\noindent The language of (\px) consists of the six trees \{$x_1$ = [\sub{S}
[\sub{A} a] [\sub{A} a]], $x_2$ = [\sub{S} [\sub{B} a a]], $x_3$ =
[\sub{S} [\sub{A} b] [\sub{A} b]], $x_4$ = [\sub{S} [\sub{B} b b]],
$x_5$ = [\sub{S} [\sub{A} a] [\sub{A} b]], $x_6$ = [\sub{S} [\sub{A}
b] [\sub{A} a]]\} with probability distribution $q: x_1 \mapsto
2/9, x_2 \mapsto 1/4, x_3 \mapsto 1/18, x_4 \mapsto 1/4, x_5 \mapsto
1/9, x_6 \mapsto 1/9$.

We sample from $q$ via stochastic derivations.  In a stochastic
derivation, we start with the start symbol, $S$.  There are two rules
expanding S: \mbox{S \goto\ A A} and \mbox{S \goto\ B}.  We flip a
coin to choose between them, heads for \mbox{A A}, tails for B.  Suppose the
coin comes up heads.  We expand S to \mbox{A A}, and then expand each of the
A's in turn.  To expand the first A, we consider the two rules \mbox{A
\goto\ a} and \mbox{A \goto\ b}.  To decide between them, we flip a
loaded coin that comes up heads (\mbox{A \goto\ a}) 2/3 of the time
and tails (\mbox{A \goto\ b}) 1/3 of the time.  Suppose this coin also
comes up heads.  We rewrite the first A as a and go to the second A.
We flip the loaded coin again; suppose it comes up heads again.  We
rewrite the second A as a, and the result is tree $x_1$.  The chances
of throwing three heads in this manner are $1/2 \cdot 2/3 \cdot 2/3 =
2/9 = q(x_1)$.  If we sample repeatedly in this manner, the proportion
of tree $x_1$ in the resulting corpus will converge to 2/9.  This is
the sense in which stochastic derivations of this sort sample from the
distribution defined by the given SCFG.

When we went from SCFGs to random fields, we lost the transparent
connection between the probability distribution defined by the field
and a method for sampling from it.  Since weights do not sum to one
for rules with the same lefthand side---indeed, since the properties
with which weights are associated are not even necessarily rule
applications---we cannot sample in the same way as we sample from an
SCFG.

There is, however, a method that can be adapted for sampling from the
random field defining a probability distribution over the language of
an AV grammar.  This method is the Metropolis-Hastings algorithm.
Specifically, in the case of sets of dags with probability
distribution $q$, we proceed as follows.  

Recall that we have a grammar $G$ consisting of feature structures.  We also have a context-free
analogue $H$ of $G$ with weights $\theta$, which we use to define the
initial distribution $p(x)$.  In addition, we have a field consisting of a set of
properties $f_i$ with weights $\beta_i$.  The grammar defines a set of
$\Omega = L(G)$ and the field plus initial distribution define a
probability distribution $q(x) = \frac{1}{Z} \prod_i \beta_i^{f_i(x)} p(x)$ over $\Omega$.

We can sample from the initial distribution $p(x)$ by
performing stochastic derivations using grammar $H$.  The derivations
map to dags in $L(G)$ according to the correspondence between
context-free rules and the AV rules of $G$.  It is possible that some
of the derivations will fail---that they will map to inconsistent
dags.  Those derivations are simply discarded.  That is, the
probability that $H$ assigns to a derivation is actually
$\breve{p}(x)$; when we throw away derivations that map to
inconsistent dags, the result is to restrict $\breve{p}(x)$ to
consistent dags and normalize it, so that we end up sampling from $p(x)$.

In this way, we can sample from $L(G)$, but not in accordance with the
field probability $q(x)$.  The essence of the Metropolis-Hastings
algorithm is a means of converting the sampler for $p(x)$ into a
sampler for $q(x)$.  Suppose we are generating a corpus, and have
generated dags $x_1, \ldots, x_n$.  Now we wish to add another dag,
$x_{n+1}$, to the corpus.  We generate a dag $y$ at random using the
sampler for $p(\cdot)$.  Now, instead of simply adding $y$ to the
corpus, we flip a loaded coin, that comes up heads with probability

\begin{mathexample}\label{mh}
A(y|x) = \min\{1,\frac{q(y)p(x_n)}{q(x_n)p(y)}\}
\end{mathexample}

\noindent If the coin comes up heads, we do include $y$ in the corpus,
that is, $x_{n+1} = y$.  But if the coin comes up tails, we throw $y$
away and make a copy of $x_n$ instead, that is, $x_{n+1} = x_n$.

The acceptance probability $A(y|x)$ reduces in our case to a
particularly simple form.  If $q(y)p(x_n) \geq q(x_n)p(y)$, then obviously
$A(y|x) = 1$.  Otherwise, writing $F(x)$ for the ``field weight'' $\prod_i
\beta_i^{f_i(x)}$, we have:

\[\begin{array}{lcl}
A(y|x) &=& \frac{Z^{-1} F(y) p(y) p(x_n)}
{Z^{-1} F(x_n) p(x_n) p(y)} \\
       &=& F(y)/F(x_n)
\end{array}\]

It can be shown that the result of generating a new dag with
probability $p(\cdot)$ and accepting it with probability
$A(\cdot|x_n)$ yields a sampler for $q(\cdot)$ (see e.g.\
Winkler \cite{Winkler_95}).  The final ``acceptance'' step intuitively serves the
role of ``punishing'' dags that the $p$-sampler proposes more
often than a $q$-sampler would, and shifting their probability to dags
that the $p$-sampler would propose less often than a $q$-sampler
would.

In somewhat more detail, if we think of the corpus $x_1, x_2, \ldots$
as a random walk through the space $L(G)$, the Metropolis-Hastings
algorithm works because it forces the random walk to spend time in a
region $R$ proportional to the probabiliy of $R$.  This is
accomplished, intuitively, by preservation of what is known as {\it
detailed balance.}  Detailed balance requires that the probability
of making a transition from dag $x$ to dag $y$
in the course of the random walk should balance the probability
of making a transition from dag $y$ to dag $x$.

Let $q(x)$ be, as always, the model probability that we wish to sample
from and let $q(y|x)$ be the transition probability---the probability
of the next dag in the corpus being $y$ if the previous dag is $x$.
In our case, $q(y|x)$ (for $y \neq x$) is the probability that we
generate $y$ at random, and then also accept it: $q(y|x) = p(y) A(y|x)$.
Define $q(x,y)$ (for $x \neq y$) to be the joint probability that $x$
is the previous dag and $y$ is the next dag; that is, $q(x,y) = q(x)
q(y|x)$.  Detailed balance requires that $q(x,y) = q(y,x)$.
If detailed balance is preserved, it can be shown that the
empirical distribution of the corpus generated by the random walk
converges to $q(\cdot)$, and that the expectation of a function $f$
taken with respect to the empirical distributions converges to $q[f]$.

We can see that the transition probability we have assumed does indeed preserve
detailed balance, as follows.  Let $x$ be the last-generated tree and
$y$ the new tree, and suppose that $q(y)p(x) > q(x)p(y)$.  Then:

\[\begin{array}{lcl@{\hspace{2em}}lcl}
q(y|x) &=& p(y)  &   q(x|y) &=& p(x) \frac{q(x)p(y)}{q(y)p(x)} \\
       &&        &          &=& \frac{q(x)}{q(y)}p(y) \\[2ex]
q(x,y) &=& q(x)p(y) &  q(y,x) &=& q(y) \frac{q(x)}{q(y)}p(y) \\
       &&           &         &=& q(x)p(y)
\end{array}\]

\noindent That is, $q(x,y) = q(y,x)$ and detailed balance is
confirmed.  The remaining cases $q(y)p(x) < q(x)p(y)$ and $q(y)p(x) =
q(x)p(y)$ are similar and are left as an exercise for the reader.

\section{Final Remarks}

In summary, we cannot simply transplant CF methods to the AV grammar
case.  In particular, the ERF method yields correct weights only for
SCFGs, not for AV grammars.  We can define a probabilistic version of
AV grammars with a correct weight-selection method by going to random
fields.  Property selection and weight adjustment can be accomplished
using the DDL algorithms.  In property selection, we need to use
random sampling to find the initial weight for a candidate property,
and in weight adjustment we need to use random sampling to solve the
weight equation.  The random sampling method that DDL used is not
appropriate for sets of dags, but we can use the Metropolis-Hastings
method.

As a closing note, it should be pointed out explicitly that the random field
techniques described here can also be profitably applied to
context-free grammars.  As Stanley Peters nicely put it, there is a
distinction between {\it possibilistic} and {\it probabilistic}
context-sensitivity.  Even if the language described by the grammar of
interest---that is, the set of possible trees---is context-free, there
may well be context-sensitive statistical dependencies.  Random fields
can be readily applied to capture such statistical dependencies
whether or not $L(G)$ is context-sensitive.




\begin{thebibliography}{1}

\bibitem{Brew_95}
Chris Brew.
\newblock Stochastic {HPSG}.
\newblock In {\em Proceedings of EACL-95}, 1995.

\bibitem{Eisele_94}
Andreas Eisele.
\newblock Towards probabilistic extensions of constraint-based grammars.
\newblock Technical Report Deliverable R1.2.B, DYANA-2, 1994.

\bibitem{Mark_92}
Kevin Mark, Michael Miller, Ulf Grenander, and Steve Abney.
\newblock Parameter estimation for constrained context-free language models.
\newblock In {\em Proceedings of the Fifth Darpa Workshop on Speech and Natural
  Language}, San Mateo, CA, 1992. Morgan Kaufman.

\bibitem{Miller_90a}
M.I. Miller and J.A. O'Sullivan.
\newblock Entropies, combinatorics and probabilities of context-free branching
  processes.
\newblock Technical report ESSRL-90-16, Electronic Systems and Signals Research
  Laboratory, Washington University, 1990.

\bibitem{DellaPietra_95}
Stephen~Della Pietra, Vincent~Della Pietra, and John Lafferty.
\newblock Inducing features of random fields.
\newblock tech report CMU-CS-95-144, CMU, 1995.

\bibitem{Riezler_96}
Stefan Riezler.
\newblock Quantitative constraint logic programming for weighted grammar
  applications.
\newblock Talk given at LACL, September 1996.

\bibitem{Winkler_95}
Gerhard Winkler.
\newblock {\em Image Analysis, Random Fields and Dynamic Monte Carlo Methods}.
\newblock Springer, 1995.

\end{thebibliography}
\end{document}